\documentclass{article}
\usepackage{kr}

\usepackage{times}
\usepackage{soul}
\usepackage{url}
\usepackage{comment}
\usepackage[hidelinks]{hyperref}
\usepackage[utf8]{inputenc}
\usepackage[small]{caption}
\usepackage{graphicx}
\usepackage{amsmath}
\usepackage{amsthm}
\usepackage{booktabs}
\usepackage{algorithm}
\usepackage{algorithmic}
\title{Online Grounding of Symbolic Planning Domains in Unknown Environments}

\author{%
Leonardo~Lamanna$^{1,2}$\and
Luciano~Serafini$^1$\and
Alessandro~Saetti$^2$\and
Alfonso~Gerevini$^2$ \and
Paolo~Traverso$^1$ 
\affiliations
$^1$Fondazione Bruno Kessler, Trento, Italy\\
$^2$University of Brescia, Italy 
\emails\tt\footnotesize
\{lamanna,serafini,traverso\}@fbk.eu 
\{alessandro.saetti,\\alfonso.gerevini\}@unibs.it
}

\usepackage[backgroundcolor=green!20,textsize=scriptsize,textwidth=4em]{todonotes}
\usepackage{bm}
\usepackage{tikz}
\usetikzlibrary{shapes}
\usepackage{caption}
\usepackage{subcaption}
\usepackage{multirow}
\usepackage{hyperref}
\usepackage{url}
\usepackage{refcount}
\usepackage{amsfonts}

\usepackage{listings}
\usepackage{verbatimbox}
\usepackage{mwe}
\usepackage{soul}
\usepackage{verbatim}
\usepackage{fancyvrb}
\usepackage{pgfgantt}
\usepackage{xcolor}
\usetikzlibrary{positioning, arrows, automata, shapes,decorations.pathreplacing,calligraphy}

\usepackage{mwe}
\usepackage{verbatimbox}

\colorlet{punct}{red!60!black}
\definecolor{background}{HTML}{EEEEEE}
\definecolor{delim}{RGB}{20,105,176}
\colorlet{numb}{magenta!60!black}

\lstdefinelanguage{json}{
    basicstyle=\normalfont\ttfamily,
    numbers=left,
    numberstyle=\scriptsize,
    stepnumber=1,
    numbersep=8pt,
    showstringspaces=false,
    breaklines=true,
    frame=lines,
    backgroundcolor=\color{background},
    literate=
     *{0}{{{\color{numb}0}}}{1}
      {1}{{{\color{numb}1}}}{1}
      {2}{{{\color{numb}2}}}{1}
      {3}{{{\color{numb}3}}}{1}
      {4}{{{\color{numb}4}}}{1}
      {5}{{{\color{numb}5}}}{1}
      {6}{{{\color{numb}6}}}{1}
      {7}{{{\color{numb}7}}}{1}
      {8}{{{\color{numb}8}}}{1}
      {9}{{{\color{numb}9}}}{1}
      {:}{{{\color{punct}{:}}}}{1}
      {,}{{{\color{punct}{,}}}}{1}
      {\{}{{{\color{delim}{\{}}}}{1}
      {\}}{{{\color{delim}{\}}}}}{1}
      {[}{{{\color{delim}{[}}}}{1}
      {]}{{{\color{delim}{]}}}}{1},
}

\newtheorem{definition}{Definition}

\def\actmodel{\mathcal{M}}

\def\opc{op(\bm c)}
\def\bx{\mathbf{x}}
\def\M{\actmodel}

\def\C{\mathcal{C}}

\def\V{\mathcal{V}}
\def\P{\mathcal{P}}

\def\OP{\mathcal{O}}

\def\PY{P_{\cal P}}
\def\RY{R_{\cal P}}

\def\bY{\bm Y}

\def\bc{\bm c}
\def\bx{\bm x}

\def\bz{\bm z}
\def\be{\bm e}
\def\pc{p(\bm c)}

\def\eff{\mathsf{eff}}
\def\prc{\mathsf{pre}}
\def\param{\mathsf{par}}

\def\opc{op(\bc)}
\def\opci#1{op_{#1}(\bc_{#1})}
\def\bc{\bm c}
\def\M{\mathcal{M}}
\def\C{\mathcal{C}}
\def\P{\mathcal{P}}
\def\G{\mathcal{G}}
\def\V{\mathcal{V}}
\def\PP{\langle\M,\C,s_0,\G\rangle}

\def\pickup{\mathsf{pickup}}
\def\gocloseto{\mathsf{goCloseTo}}
\def\belstate{\langle\C,\bz_\C,s,\bz_s,\mathbf{Pr}\rangle}

\def\arch{\textsc{Ogamus}}
\def\robothor{\textsc{RoboThor}}
\def\ai2thor{\textsc{iThor}}

\makeatother
\begin{document}

\maketitle

\begin{abstract}
  If a robotic agent want to exploit symbolic planning techniques
  to achieve some goal, it must be able to properly
  ground an abstract planning domain in the environment in which it operates. 
  However, if the environment is initially unknown by the agent, 
  the agent need to explore it and discover the salient aspects of the environment
  needed to reach its goals. Namely, the agent has to discover: (i) the objects present in
  the environment, (ii) the properties of these objects and their relations, and
  finally (iii) how abstract actions can be successfully executed. 
  The paper proposes a framework that
  aims to accomplish the aforementioned perspective for an
  agent that perceives the environment partially and subjectively, through
  real value sensors (e.g., GPS, and on-board camera) and can operate
  in the environment through low level actuators (e.g., move forward of 20 cm).
  %
  We evaluate the proposed architecture in photo-realistic simulated environments,
  where the sensors are RGB-D on-board camera, GPS and compass, and low level actions
  include movements, grasping/releasing objects, and manipulating objects.
  The agent is placed in an unknown environment and asked to find objects of a certain type, place an object on top of another, close or open an object of a certain type. We compare our approach with the state of the art methods on object goal
  navigation based on reinforcement learning, showing better
  performances. \end{abstract}



\section{Introduction}

Symbolic planners are powerful and flexible tools
  \cite{DBLP:books/cu/GNT2016} that, given a general symbolic description
  of an available set of actions (i.e., a planning
  domain) and a detailed description of an environment, are
  capable of generating plans for achieving ideally any goal 
  about (known) objects in the environment.
%
In several applications, the information about the environment required to instantiate a planning domain is not available from the beginning. 
In particular, when an agent is placed in a 
new environment, it does not know the objects that populate the
environment, and therefore it does not know their
specific properties and relations. 
Consider, for instance, a robot that has to move
around and manipulate objects in a kitchen (tables, chairs, apples,
etc.) without knowing which and how many objects are really in the room.
%
In this setting, the exploitation of a planning domain is a
compelling challenge for three main reasons.
First, in realistic environments, it is unfeasible for the robot to
acquire a complete/correct and sufficiently detailed description of
the environment before starting to plan and execute actions
towards the achievement of its goals.
Second, a robotic agent usually has a first-person perspective and partial view of the
environment (e.g., by an on-board camera), so that the only way to acquire symbolic knowledge suitable for planning is by
executing actions, observing their effects through its sensors, and
mapping the sensory data (e.g., raw images) in a symbolic
state.
Third, high level actions of the planning domain are not directly executable by the robot, and therefore they need to be compiled to low level actions
executable in the environment by the agent actuators.
For instance,  
the action $\gocloseto(\mathsf{table}_0)$ is compiled into a sequence of robot movements and rotations, which follow the path
provided by a path-planner, and moves the robot to the (nearest) location close to object $\mathsf{table}_0$. 

This paper proposes a framework for agents that 
incrementally instantiate a planning domain, specified in PDDL, by planning, acting,
and sensing, in an unknown environment. 
%
%
At each time-point, the belief of the agent about the current state of the
environment is represented by three components, namely:
$(i)$ the set of objects currently known by the agent and their
properties expressed with the predicates of the PDDL domain, e.g., $\mathsf{apple}_1$, $\mathsf{table}_0$, and
$\mathsf{on}(\mathsf{apple}_1,\mathsf{table}_0)$;
 $(ii)$
for each known object, a set of low-level features as perceived by the
agent, e.g., visual features and positions of $\mathsf{apple}_1$ and
$\mathsf{table}_0$;
$(iii)$ a set of global features associated to the current environment
state, e.g., an occupancy map of the environment, and
the current pose of the robot.


For this framework, we propose an online iterative algorithm, called \arch\ (Online
Grounding of Action Models in Unknown Situations), that allows an
agent, equipped with a lifted PDDL planning domain, and placed in an
unknown environment, to achieve a set of goals expressed in the
language of its PDDL action model. 
%
The agent is
  initialized without
  prior knowledge about the environment where it has to operate,
  i.e., with 
  the empty set of objects, the empty set of their properties, and all the points of the occupancy
  map set as traversable. \arch\ attempts to achieve the goal by combining four main
activities, namely: (i) \emph{exploring the environment} to acquire the
knowledge needed to achieve the goals; (ii) \emph{abstracting the
sensor information} obtained at every step into a symbolic state; (iii)
performing \emph{symbolic planning} in the abstract model grounded
with the current beliefs and current abstract state; (iv)
\emph{executing} the planned abstract actions by compiling them into low-level operations suitable for the current state of the environment.



The main features of \arch\ are the following.
\emph{Generality:} \arch\ is able to deal with any goal 
that can be expressed by a (first-order) formula using the predicates of the
PDDL domain. For instance the goal of putting ``two apples on a table'' 
can be specified by the formula 
$\exists x\, y\, z.\mathsf{on}(x,z)\wedge \mathsf{on}(y,z)\wedge
\mathsf{apple}(x)\wedge 
\mathsf{apple}(y)\wedge
\mathsf{table}(z)\wedge
x\neq y$.
Notice that goals are expressed with existentially
  quantified variables; this is because, initially, the agent is not aware
  of any objects in the domain. An important step
  necessary to achieve the goal is indeed discovering the objects for 
  replacing the existential variables.
%
%
\emph{Explainability}: The behaviour of
the agent, its plans, and the effects of actions are represented
at a symbolic level in which the states of the PDDL domain are
derived at every step by abstracting the sensory data.
%
%
\emph{Robustness:}
The action model, the obtained symbolic state representation, and the action compilation are
not required to be fault free. As experimentally shown in this paper, \arch\ achieves high
success rate even with low precision object detectors and classifiers. 
%


We have implemented and experimentally evaluated \arch\ 
in the \ai2thor\ \cite{ai2thor} and \robothor\ \cite{robothor}
simulated photo-realistic environments for embodied AI.  We evaluate
\arch\ on different tasks including 
``go close to an object of a certain type'' (e.g., ``go close to a
fridge'').  In the area of Embodied AI,
this class of tasks is called ``object goal
  navigation'', and it is still considered a challenging and open problem. 
  Our 
  framework and evaluation go beyond such a challenge, considering more complex
  tasks, such as ``open/close an object of a certain type''
(e.g., ``open a laptop''), and ``put an object of a type on top of an
object of another type'' (e.g., ``put an apple on a table'').
We compared our results with the state-of-the-art
  approaches on the object goal navigation task. In particular, we
  show that, in the benchmarks of the last \robothor\
  object goal navigation challenge, our approach outperforms the state
  of the art methods based on Reinforcement Learning
  (RL).\footnote{\href{https://ai2thor.allenai.org/}{https://ai2thor.allenai.org/}}
  The evaluation on the other tasks leads to very positive
  performance, but we cannot compare them with any state-of-the-art
  approach since, to the best of our knowledge, we are the first
  solving such types of tasks in the \ai2thor\ and \robothor\
  environment.

The paper is structured as follows: firstly we analyze the related
literature, then we describe the framework
and the algorithm adopted by the agent to reach a specific goal in an
unknown environment; finally, we present the experimental evaluation
and a comparison with RL-based approaches.




\section{Related Work}

The problem of integrating symbolic action models with low level sensory data and actions has been addressed by different approaches. Most of them are based on RL techniques.
\citeauthor{lyu2019sdrl} (\citeyear{lyu2019sdrl}) propose a framework, called SDRL, which combines
symbolic planning on PDDL and Deep RL to learn policies that compile
high level actions into low level operations. SDRL assumes that a
grounded domain model is provided in input and never
updated. A fundamental difference w.r.t.\ our approach is that \arch, instead, learns how to ground the domain model with new
objects discovered online. Moreover, SDRL assumes a perfect
oracle that maps low level perceptions into symbolic states, while
\arch\ deals with faulty mappings without assuming to have an oracle.

NSRL \cite{ma2021learning} represents abstract domains in first-order logic and uses RL to learn high level policies. NSRL generates
a compact representation of the learned policies as a set of rules via Inductive Logic
Programming. Similarly to SDRL, NSRL assumes a given and fixed
abstract domain instantiation and a perfect mapping from sensory data to
symbolic states, while \arch\ overcomes these assumptions.
 
DPDL \cite{kase2020transferable} represents abstract domains in 
PDDL. It learns online both mappings from sensory data to symbolic states
and low level policies for high level actions. In \arch, instead, the mapping
from perceptions to symbolic states is obtained
by combining a set of neural networks that are trained off-line.
Moreover some of the high
level actions are pre-compiled in low level operations (e.g., pick-up
an object at a given position), while policies for moving actions are computed online
via path planning. As the other methods mentioned above, DPDL assumes a given and fixed
grounded PDDL domain, while this is not the case for \arch. Moreover, it focuses on performing 
manipulation tasks in a single scene type (i.e., a kitchen). \arch, instead, can work on 
different scenes (we evaluate it on 35 different scenes grouped by kitchens, living-rooms, bedrooms, bathrooms and apartments). Finally, DPDL works with an external fixed camera, while \arch\ uses egocentric and dynamic views. This makes the tasks more difficult for \arch, since the agent needs to navigate, explore the environment, and find new objects outside of its current view.

Differently from the above mentioned works, in the approach proposed by
\citeauthor{garnelo2016towards} (\citeyear{garnelo2016towards}), like
in \arch, the
agent instantiate the abstract domain online by 
augmenting the set of objects every time it discovers new ones.
The states of the instantiated abstract model is
represented with a set of propositional atoms on the current set of constants.
However, this approach is evaluated
 only with an extremely simple environment, while \arch\ is tested in
accurate and photo-realistic simulated environments with complex objects and egocentric views. Furthermore, the approach in
\cite{garnelo2016towards} does not take advantage of the power of
symbolic planning techniques on PDDL domain descriptions, and it
does not generalize over different tasks. On the contrary, \arch\ exploits symbolic planning which allows to accomplish variegated tasks that can be specified using PDDL.

We experimentally evaluate \arch\ on the Object Goal Navigation task
that has recently received much attention in the embodied AI
community
\cite{mirowski2017learning,savva2017minos,fang2019scene,mousavian2019visual,campari2020exploiting,wortsman2019learning,chaplot2020learning,chaplot2020object,ye2021auxiliary}.
We experimentally show that, in the benchmark of the last  \robothor\ object goal navigation challenge, \arch\ performs better than 
a method based on DD-PPO \cite{wijmans2019ddppo}, which won the challenge using pure RL
based on low-level features, without exploiting a symbolic domain. 

%

\section{The Framework}

In this section, we start by introducing the basic definitions of our reference 
framework. We then introduce the \arch\ algorithm. The section closes with
the description of what are the basic components that need to be
specified in order to cope with new abstract predicates or actions
of the PDDL domain. 

\subsection{Preliminary definitions} 
Let $\P$ be a set of first order predicates, 
$\V$ a set of variables (also called parameters),
and  $\C$  a set of constants.
We use $\P(\V)$ to denote the set of
atoms $P(x_1,\dots,x_m)$, where $x_i\in\V$ and $P\in\P$,
and
$\P(\C)$ to denote
the set of atoms obtained by grounding $\P(\V)$ with the constants in
$\C$.

\begin{definition}[Action model]
  Given a set of operators $\OP$,
  an \emph{action model} $\M$ associates to each $op\in\OP$
  an \emph{action schema}, which is a tuple
  $\left<\param (op),\prc(op),\eff^+(op),\eff^-(op)\right>$, 
  where $\param(op)\subseteq\V$, 
  $\prc(op)$, $\eff^+(op)$, and $\eff^-(op)$ are subsets of $\P(\param(op))$.
\end{definition}

\begin{definition}[Ground action]
  The \emph{ground action} $\opc$ of an operator $op\in \OP$
  with $\bc = \langle c_1,\dots,c_n \rangle$ constants in $\C$ is the 
  tuple $\langle \prc(\opc),\eff^+(\opc),\eff^-(\opc)\rangle$,
  obtained by instantiating the atoms of $\prc(op)$, $\eff^+(op)$, and $\eff^-(op)$ with~$\bc$.
\end{definition}

\begin{definition}[Planning problem]
  A \emph{planning problem} is a tuple $\PP$ where $\M$ is an action
  model, $\C$ is a (possibly empty) set of constants, $s_0\subseteq\P(\C)$ is the
  initial state, and $\G$ is a
  first order formula over 
  $\P$, $\V$ and $\C$.
\end{definition}


\begin{definition}[Plan]
  A \emph{plan} for a planning problem $\PP$ is a sequence
  $\langle \opci1,\dots,\opci{n} \rangle$ such that there is a sequence
  $\langle s_1,\dots,s_n \rangle$ of subsets of $\P(\C)$ (aka states),
  such that for every $0\leq i < n$, $\prc(\opci{i})\subseteq s_i$, 
  $s_{i} = s_{i-1}\cup\eff^+(\opci{i})\setminus\eff^-(\opci{i})$,  and 
  $s_{n}\models\G$.
\end{definition}

\noindent
Notice that our definition of planning problem allows to express first-order goal formula $\G$. We say that a state $s\models\G$ iff $\bigwedge_{P(\bc)\in
      s}P(\bc)\wedge\bigwedge_{P(\bc)\in \P(\C)\setminus s}\neg
    P(\bc)\models\G$, under the assumption that all the elements of
    the problem are in $\C$.


In order to use an abstract model, 
an agent needs to \emph{anchor} the symbols occurring in the states of
the planning domain with the real-world perceptions, and to map abstract
actions into actions executable in the real world \cite{coradeschi2003introduction}.
We suppose that the agent can \emph{partially} observe the current state of
the environment through a set of sensors, for instance images provided by an
RGB-D camera, which do not directly correspond to the states of the abstract
model. Furthermore, the set of sensors provide only a 
partial and subjective view of the environment. For instance, the RGB-D camera
provides only an egocentric view of a portion of the room visible by
the agent.
We also suppose that the agent interacts with the environment by executing low-level operations
(e.g., move 25 cm forward, rotate $30^\circ$ left, pick up or put
down an object at the GPS-coordinates $(x,y,z)$), which are different
from the actions in the abstract action model. 
%
We need therefore to link the abstract state to real perceptions, and
the abstract actions to operations executable by the actuators of the agent.
Let us first consider the relationship between abstract states and
perceptions. 

\begin{description}
  \item[Object and state anchoring.]
Every object that the agent is aware of at a given instant 
is represented by a constant $c\in\C$ that is the internal identifier for such an
object.  Following the approaches to symbol anchoring proposed in the
literature \cite{coradeschi2003introduction,persson2019semantic},
every constant $c\in\C$ is associated with a tuple of numeric features
denoted by $\bz_c$. For instance, $\bz_c$ might include the estimated position of $c$ and a set of visual features of the different
views of $c$.  
In addition, for each state $s$ determined by the agent, we have a vector of 
state features
$\bz_s$, consisting of the 3D position of the agent in the environment, the orientation of the agent relative to its initial pose, the information about the success of the last low-level operation made by the agent, and an occupancy map of the environment.
The occupancy map is a 2D map of the environment
   storing the areas that are believed to be traversable by the
  robot. The occupancy map is initialized so that every point is traversable.
\item[Predicate predictors.] In order to map the perceptions about
   objects into atoms of the symbolic state, the agent
  associates to every predicate a probabilistic model, e.g., a neural network,
  that computes the probability of a certain atom $P(\bc)$ to be true given the features associated
  to $\bc$ and the current state ones, i.e., $Pr(Y_{P(\bc)}=\mathrm{True}\mid\bz_{\bc},\bz_s)$,
   where
  $Y_{P(\bc)}$ is a boolean random variable associated to the atom
  $P(\bc)$.
  These probabilistic
  models can be updated during execution on the basis of
    new observations. In this paper, however, we suppose
  that these probabilistic models are given (e.g., a pre-trained neural network), 
  and they are not modified during execution. 
\end{description}
We call {\em belief state} the agent's knowledge about object/state anchoring and predicate predictors.
\begin{definition}
  An agent belief state is a 5-tuple $\belstate$ where:
  \begin{itemize}
  \item $\C$ is a set of constants;
  \item $\bz_{\C}=\{\bz_c\}_{c\in\C}$ is a set of object feature vectors $\bz_c$; 
  \item $s\subseteq \P(\C)$ is the set of atoms that are believed to be true;
  \item $\bz_s$ is a vector of state features;
  \item $\mathbf{Pr}=\{Pr(Y_{P(\bc)}\mid\bz_{\bc}, \bz_{s})\}_{P\in \P}$ 
    is the set of probabilistic models used to predict
  the truth value of $P(\bc)$ given the features $\bz_{s}$ and $\bz_{\bc}$ associated with the
  constants in $\bc$.
  \end{itemize}
\end{definition}

\subsection{The \arch\ algorithm}

So far, we have not considered how the set $\C$ of constants
identifying objects is obtained by the agent.  We do not assume that
they are given a priori to the agent; instead, we are interested in
providing the agent with the capability to discover objects by adding new constants to the representation of the environment, updating
the anchor to an object, merging two constants anchored to the same
object, and deleting a constant from the representation that was erroneously identifying a non existing object in the environment.
%

%
Let $\bx$ be the
vector that contains the data returned by the sensors (i.e., the
observations) at a given time; the agent extracts from $\bx$
a set of objects $\C_{\bx}$, and for each object
$c\in\C_{\bx}$ a feature vector $\bz_c$. Since the agent can also
recognize objects that it has already seen, it is possible that
$\C_{\bx}\cap\C\neq\emptyset$. 
%


\begin{algorithm}[ht]
\caption{\arch\ algorithm}
\label{alg:algorithm}
\textbf{Input}: $\actmodel$, $\G$, $\mathbf{Pr}$ and $\textsc{MaxIter}\in\mathbb{N}$.\\
\textbf{Output}: \textsc{Success}/\textsc{Fail}
\begin{algorithmic}[1]
\STATE $\langle\C,\bz_\C,s,\bz_s\rangle\gets\langle\emptyset,\emptyset,\emptyset,(\mathbf{0},nil,\emptyset)\rangle$ \label{algline:initialization}
\FOR {$1=0$ \TO \textsc{MaxIter}}
\IF {$s\models\G$} \RETURN \textsc{Success} \label{algline:success}
\ENDIF
\STATE $\pi\gets\textsc{Plan}(\actmodel,\C,s,\G)$ \label{algline:plan}
\IF{$\pi = \textsc{None}$}
\STATE $\be\gets \textsc{explore}(\bz_s)$ \label{algline:explore}
\ELSE
\STATE $\opc\gets\textsc{Pop}(\pi)$
\STATE $\be \gets \textsc{compile}(\opc,\bz_{\bc},\bz_s)$ \label{algline:compile-plan}
\ENDIF
\STATE $e_1 \gets${\it Pop}$(\be)$
\STATE $\bx\gets \textsc{Exec}(e_1)$ \label{algline:exec}
\STATE $\bz_s \gets \textsc{getStateFeatures}(\bx)$ \label{algline:get-stateFeatures}
\STATE $\C_{\bx},\bz_{\C_{\bx}} \gets \textsc{GetObjs}(\bx)$ \label{algline:get-objects}
\STATE $\C,\bz_\C \gets \textsc{updateObjs}(\C,\bz_\C,\C_{\bx},\bz_{\C_{\bx}})$ \label{algline:merge-objects}
\STATE $\textit{Pr}(\bY_{\P(\C)}) \gets \textsc{PredictState}(\bz_{\C},s,\bz_s)$ \label{algline:predict-state}
\STATE $s \gets \{\pc\in\P(\C) \mid
\textit{Pr}(Y_{\pc}=\textit{True}\mid \bz_{\bc})>\!1\!-\!\epsilon\}$ \label{algline:decide-state}
\IF{$\pi {\not=}$ \textsc{None} and \textsc{succeed($\opc$)}}
\STATE $s \gets s \cup \eff^+(\opc) \setminus \eff^-(\opc))$ \label{algline:update-state}
\ENDIF
\ENDFOR
\RETURN \textsc{Fail} \label{algline:fail}
\end{algorithmic}
\end{algorithm}

In the following, we shortly describe the \arch\ algorithm (Algorithm \ref{alg:algorithm}).

\begin{itemize}
\item The algorithm takes as input an action model $\actmodel$,
  a set $\mathbf{Pr}$ of probabilistic models for
    predicting the predicates in $\P$, a goal formula $\G$, and a
  maximum number of iterations.
Notice that the goal $\G$ cannot contain constants, since we suppose
that at the beginning the agent is not aware of any object. For instance, the goal requiring that an apple is inside a box can be encoded by the PDDL expression representing formula $\exists x,y~\mathsf{apple}(x) \land \mathsf{box}(y) \land \mathsf{in}(x,y)$.
  The main objective of the algorithm is to reach a state that achieves the goal.
\item The agent starts by initializing all the components of its
   state to the emptyset (line \ref{algline:initialization}). We assume indeed that the
  agent is not aware of any object in the environment,
  therefore $\C=\emptyset$. Since $\C$ is
  empty, $\bz_C$, $\P(\C)$ and $s$ are also empty. The information in $\bz_s$ 
  representing the position and orientation of the agent is
  initialized with a vector of $0$'s; 
  the information in $\bz_s$ about the success of the last
  operation is set to
  {\em nil}; finally, the occupancy map of the environment in $\bz_s$ is
   set to an empty map so that all the points are traversable.
\item Then the agent iterates for a maximum number of steps,
  checking if the current state $s$ satisfies the goal (line
  \ref{algline:success}); when this is the case, it returns \textsc{Success}.
\item Otherwise, the agent invokes a planner (line \ref{algline:plan})
  to solve the planning problem defined on the input action model, the current set of
  objects, the current state, and the input goal formula $\G$.
  
\item If the planner does not find a plan that satisfies the goal, then the agent
  explores the environment in order to discover new objects that are needed to satisfy the goal.
  For instance, if the goal is to put an apple into a box, then the planner
  can find a plan only if in the current state $s$ there is at least one object of type
  apple and one of type box. 
For the exploration phase (line \ref{algline:explore}), the agent
  randomly selects a target position on the occupancy map (stored in
  $\bz_s$) that it believes to be free from other obstacles.
  As shown in Figure~\ref{fig:example_path}, $\textsc{explore}(\bz_s)$
  calls a path planner that checks if such a position is reachable (if
  it is not reachable a new position is selected) and returns a
  sequence $\bm e$ of low-level navigation and rotation operations,
  which, according to the current
  knowledge of the agent, moves the agent from its current position to
  the selected target. For efficiency reasons, this path
    is computed in an approximated 
  occupancy map obtained by discretizing the occupancy map through a grid.
  \begin{figure}
      \centering
      \input{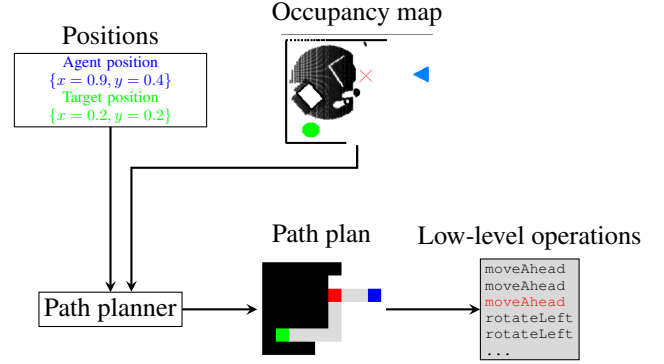}
      \caption{Example of \textsc{explore($\bz_s$)}. The occupancy map, agent position (in blue) and target position (in green) are given as input to a path planner which discretizes the occupancy map, computes a path plan, and compile the path plan into a sequence of low-level operations. }
      \label{fig:example_path}
  \end{figure}
    \begin{figure*}
      \centering
      \input{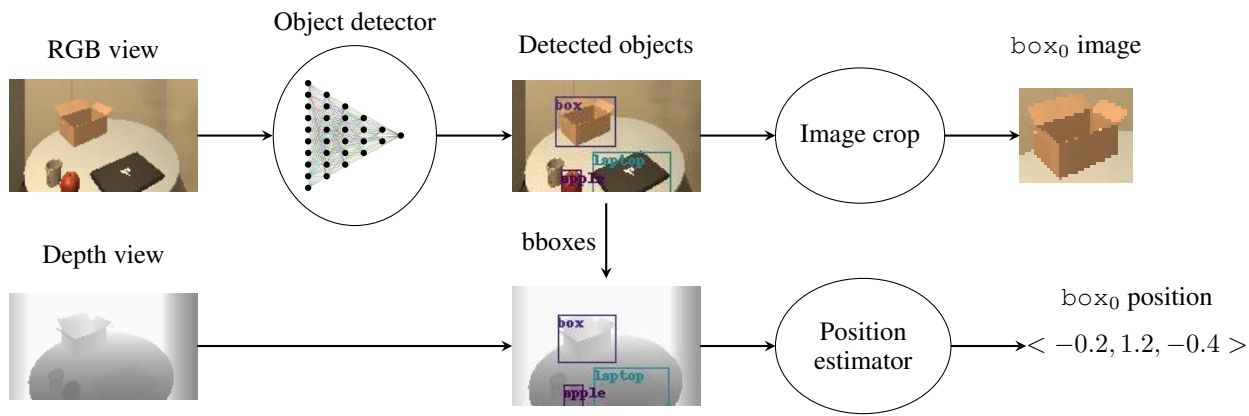}
      \caption{Example of object features extraction. The object detector takes as input the perception composed by the agent view RGB image, and returns a set of bounding boxes together with the detected object types. The object bounding boxes are used with the agent view depth image to estimate their positions (in meters w.r.t.\ the initial agent position). The lighter the pixels in the box, the farther the associated object.}
      \label{fig:example_object_detection}
  \end{figure*}
  The execution of such a sequence of operations might fail due to the
  partial or incorrect knowledge of the agent, i.e.,
    when the agent wrongly believes that a certain area on the path returned by the path-planner is traversable while there is an obstacle. In Figure \ref{fig:example_path}, the agent fails to reach the red cell. Indeed, the execution of the third $\mathsf{moveAhead}$ action fails because the robot bumps into a table, whose size was only partially determined at the beginning of the exploration phase.   
  The exploration terminates whenever the goal is reachable according to the learned problem. For example, consider the goal requiring that an apple is in a box and assume that an apple has already been discovered in the environment. The execution of the computed sequence of operations terminates in advance, when the agent detects the presence of a box on the table in Figure \ref{fig:example_path}, as it approaches the table by executing the first operations in the sequence. 
\item If instead the planner succeeds and returns a valid plan $\pi$,
  then the first action of $\pi$ is compiled into a sequence of
  low-level operations $\bm e$ (line \ref{algline:compile-plan}). The compilation of
  the action is based on the object and state features available in the
  agent's  state. For instance,  the high level action
  $\pickup(c)$ is compiled into the low-level operation $\pickup(x,y,z)$
  where $(x,y,z)$ is the current (believed) position of object $c$, memorized in $\bz_{c}$. The simulator executes such an operation by picking up what is present at these coordinates.
To compile the action $\gocloseto(c)$ instead, the agent calls a path-planner that provides
a path from the current position of the agent (memorized in
$\bz_s$) to a position close to $c$. 
\item Successively, the first operation of sequence $\bm e$ is executed 
  (line \ref{algline:exec}), and a new observation $\bm x$ is
  obtained. The execution of the first operation may
    fail or not. In both cases, the agent can acquire new knowledge
    (e.g., discover new objects or an obstacle), which can be used to
    produce a better compilation of an high-level action, and/or
    produce a better plan. 
  %
  %
  Then, the new state features $\bz_s$ are
  extracted from the sensory data $\bx$
  (line \ref{algline:get-stateFeatures}).
  The information about the 
  occupancy map is updated using the
  information of success/failure of the action and the depth image.
  
\item Then the agent runs an object detector (line
  \ref{algline:get-objects}) on the RGB image contained in observation
  $\bx$ which  returns a set of objects $\C_{\bx}$, each associated
  with a vector of numeric features $\bz_{c}$. 
  These features include the
  bounding box, an estimation of the object position, and a vector of
  visual features extracted from the cropping of the
  image with the bounding box. Figure \ref{fig:example_object_detection} gives an example of extraction of a number of objects, including a box, from the egocentric view of an agent robot, together with the bounding-box image of such a box and the estimate of its position.
%
%

%
\item Next, at line \ref{algline:merge-objects}, the agent merges the
  objects $\C_{\bx}$ recognized in the current perception with the
  ones already known, i.e., $\C$. For every object ${c'}\in\C_{\bx}$
  there are two possible situations: $(i)$ ${c'}$ does not match with
  any object $c\in\C$, and therefore it is added to $\C$ with the
  corresponding features $\bz_{{c'}}$; $(ii)$ ${c'}$ matches with a
  $c\in\C$; in this case the features $\bz_{c}$ of $c$ are
  extended/updated with the features $\bz_{c'}$.  In the
  implementation, we use a very simple matching criteria which
  considers only the estimated position of the objects. Two objects are matched when 
  their distance is less then a given threshold (set to 20cm). More
  sophisticated criteria can be adopted by defining a suitable
  distance measure between the entire set of object features. However,
  this simple criteria turned out to be sufficiently effective in our
  experiments.
      
\item In line \ref{algline:predict-state}, the agent predicts the
  truth values of each atom in $\P(\C)$ for the updated set of
  constants $\C$ by applying the predictors $\mathbf{Pr}$ on the
  features $\bz_\C$. For predicate $\mathsf{closeToAgent}$, the prediction takes also as input the agent position in $\bz_s$.
  All the atoms involving new or merged
  objects must be evaluated; the remaining atoms are evaluated
  only if the corresponding predictor takes as input some feature that
  has been updated after the execution of the last action.
  For instance, if the agent executes a move action, then all
  the atoms $\mathsf{closeToAgent}(c)$ for all $c\in\C$ must be evaluated. 
  Each atom $\mathsf{closeToAgent}(c)$ is predicted
    true if the euclidean distance between the position of the object
    represented by $c$ in $z_c$ and the agent position in $z_s$ is
    lower than a given threshold (set to $140$ cm).
  When the action $\mathsf{open(box_0)}$ is executed,  
the visual features of $\mathsf{box_0}$ probably change, and 
the truth value of predicate $\mathsf{isOpen}(\mathsf{box_0})$ is predicted as depicted in Figure \ref{fig:example_predicate_open}.
Notice that, after executing an open operation it is not guaranteed
that the object will be open, as the action might fail.

  \begin{figure}
      \centering
      \input{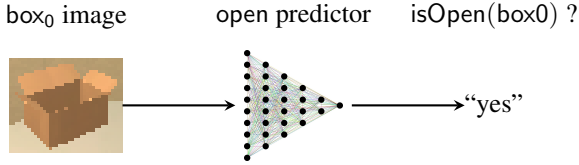}
      \caption{Example of predictor for the $\mathsf{open}$ predicate. The predictor takes as input the RGB images associated to $\mathsf{box_0}$, and returns the predicted truth value of $\mathsf{isOpen}(\mathsf{box_0})$.}
      \label{fig:example_predicate_open}
  \end{figure}
\item At line \ref{algline:decide-state}, the new state $s$ is created
  with all predicates $P(\bc)$ such that
  $\textit{Pr}(Y_{P(\bc)}=\textit{True}\mid \bz_{\bc},\bz_s)$ is higher than a given
  threshold $1 - \epsilon$ with $\epsilon \in [0,1]$.
  Our approach does not assume to have access to the correct abstract state. Indeed, the agent can produce inconsistent states (e.g., a box is both on the table and on another box), or states that do not comply with action effects. Inconsistent states do not prevent \arch\ to further plan and, whether a failure occurs, revise the agent's knowledge making them consistent. In the second case, the agent monitors the execution of high-level actions by comparing the state predicted by the PDDL model with the perceived state,
  and it solves possible inconsistencies in favour of the action effects in the model. 

\item At line \ref{algline:update-state}, when \textsc{succeed}$(\opc)$ is true, i.e., the entire sequence $\be$ of operations compiling the first action $\opc$ of $\pi$ is successfully executed, the state $s$ is updated according to the effects  of $\opc$.

\item If the agent does not reach a state $s$ that satisfies the goal $\G$
  after \textsc{MaxIter} steps, then the algorithm returns \textsc{Fail} (line \ref{algline:fail}).
\end{itemize}

\subsection{Knowledge revision for new tasks}
To show the generality and the modularity of the proposed framework, we
describe how it can be easily extended to cope with new tasks that can
possibly involve a set of new (PDDL) actions, predicates, and object
types.
To allow the agent to accomplish a new task $t_{new}$, we firstly need
to encode $t_{new}$ in a PDDL goal formula. If the encoding of
$t_{new}$ does not require the introduction of new predicates,
actions, or object types, then to solve the task it is sufficient to
invoke \arch\ with the goal formula encoding $t_{new}$.  If, instead,
the encoding of $t_{new}$ requires some new predicates, actions, or
object types, then we have to provide the agent with the capability of
(1) recognizing objects of the new types, (2) predicting the truth
value of the new predicates, and (3) compiling the new actions in low-level operations executable by the agent's actuators.
Consider, for instance, the case where $t_{new}$ is the task turning a
lamp on: $t_{new}$ can be specified by the goal formula
$\exists x.\mathsf{lamp}(x)\wedge \mathsf{turned\_on}(x)$, where
$\mathsf{lamp}$ is a new object type and 
$\mathsf{turned\_on}$ a new unary predicate.

%
To detect objects with the newly introduced type 
e.g., $\mathsf{lamp}$, the object detector need to be extended and
retrained with the new object type. This implies that the upper part
of the detector (which is responsible of classifying the objects in their types) need to be 
extended with the new type and re-trained on a dataset
containing also examples for this new type.  The introduction of a new
predicate, in our example $\mathsf{turned\_on}$, requires the
deployment of a new classifier that predicts if the
predicate holds for the objects detected from the sensory data.
In case the predictor is based on a supervised learning model, then
a training dataset with object labelled with positive and negative
examples of the predicate need to be provided. 
%
%
Finally, if the new task requires adding new actions to the
PDDL model, (e.g., to make the predicate $\mathsf{turned\_on}(x)$ true/false
we need to introduce two new action $\mathsf{turn\_on}$ and $\mathsf{turn\_off}$)
we need to specify how the new action can be compiled into a sequence of
low-level operations executable by the agent. For example,
action $\mathsf{turn\_on}(c)$ will be compiled into a low-level
operation $\mathsf{turn\_on}(x,y,z)$ where $(x, y, z)$ is the believed
position of object $c$.  

\section{Experiments}

We perform two sets of experiments. First, we experimentally evaluate \arch\ with a simulated environment on four tasks that involves
going close and move objects present in a number of rooms. Then, we compare
\arch\ with a state-of-the-art approach on the specific task of object goal navigation in different apartments. 

\subsection{Evaluating \arch}
The tasks and the
corresponding goals on which we evaluate \arch\  are the following:
\begin{enumerate}
\item Object goal navigation (\textsc{objNav} $t_1$): given an object type 
  $t_1$, the agent has to find, go close to, and look at an object of type $t_1$. 
  For instance, the agent has to go close to an apple and look at it. The corresponding goal is
    $
    \exists x (\mathsf{Apple}(x) \wedge \mathsf{CloseToAgent}(x)\wedge\mathsf{Visible}(x))
    $. The agent is close to an object when
    the distance from the object is less than 1.5 meter. 
\item Open/close an object (\textsc{Open}/\textsc{Close} $t_1$): the agent is required to
  go close to an object of type $t_1$, look at it, and open/close it. 
  For instance the agent has to open a drawer; 
   the corresponding goal is
    $
    \exists x (\mathsf{Drawer}(x) \wedge \mathsf{Open}(x))
    $.
    In order to manipulate an object
  the agent need to be at a distance less than 1.5 meter.
\item Stack an object of type $t_1$ on an object of type $t_2$ (\textsc{On} $t_1$ $t_2$): the agent
  has to find two objects of types $t_1$ and $t_2$ and put the one of type $t_1$ on top of the other of type $t_2$. For instance
  the agent has to put an apple on a table. The corresponding goal is:
  $
  \exists x y (\mathsf{Apple}(x)\wedge \mathsf{Table}(y) \wedge \mathsf{On}(x,y))
  $.
\end{enumerate}

\noindent
Since these tasks do not involve numerical resources or temporal constraints, we adopted propositional PDDL planning.

\paragraph{Simulator.}
We used the \ai2thor\  \cite{ai2thor} simulator, an open source
interactive environment for Embodied AI. \ai2thor\ provides 120
different scenes, such as kitchens, living-rooms, bathrooms, and
bedrooms, and allows a realistic simulation of the environment, including the physics of the objects. The scenes contain objects of 118 different types.
The agent perceives the current state of the environment through an
RGB-D on-board camera that provides a photo-realistic rendering of the egocentric view of the agent.
The agent also perceives its relative position and orientation via a GPS
and a compass (relative to the starting pose, which is unknown). The agent can navigate the environment by moving ahead of a
given distance (set to 25cm), turn left or right, and look up or down of a
given angle (set to $30^\circ$).%
\footnote{These settings are those
indicated by the simulator developers for their proposed challenges.} The agent
can pick up objects, move them around, and change their state (e.g., a fridge
can be opened or a laptop switched on). 

For the object goal navigation task, we also considered a second simulator, \robothor\ \cite{robothor}.  
\robothor\ is another simulation environment designed to develop embodied AI agents. 
Recently, \robothor\ hosted a competition that tackles an object goal navigation challenge; in our experiments, we also compared \arch\ with the approaches that took part in the competition.



\paragraph{Object detector.}
As an object detector we used the Faster-RCNN model available in
PyTorch 1.9 \cite{paszke2019pytorch}, pre-trained on the COCO dataset
\cite{lin2014microsoft} and fine-tuned on a self-generated dataset.
In addition to the bounding box of the detected
object, the object detector returns also the classification in one of the 118 classes.
The object detector has been trained on a dataset composed by 69,095
training and validation images. 
The labeling of the dataset has been done by using the ground-truth provided by \ai2thor.
We tested it on 12,892 images obtaining a
precision and recall of 50.99\% and 65.18\%, respectively.

\paragraph{Predicate predictors.}
For predicting predicate \textsc{On}, we trained a feed-forward neural
network \cite{svozil1997introduction} with 244 input features
composed by the bounding boxes coordinates of the two objects involved in the predicate relation and the 
1-hot encoding of the two predicted classes returned by the object
detector. For such a predicate, the training (and validation) sets  
is composed of 36,344 labelled pairs of objects. 
We evaluate the prediction of predicate \textsc{On} on a test set composed  of 8678 object pairs,
obtaining 98.32\% of both precision and recall. 
For predicting the unary predicate \textsc{Open}, we used a ResNet50 neural network
\cite{he2016deep} to extract features from the cropped object
image, followed by a linear layer with input size 2048.%
\footnote{Further technical details about the hyper-parameters and
  datasets are available in the supplementary material.}
We trained it on 48,476 labelled examples, and test it on
9685 examples, obtaining 92.84\% precision and 92.54\% recall. 
The unary predicate \textsc{closeToAgent}, meaning that the agent 
is near to the object mentioned by the predicate, is computed directly from the features 
of the object. Specifically, we check if the distance between the agent position memorized in $\bz_s$  and the
object position memorized in the object feature vector is less than the manipulation distance, which is set to
$1.5$ meter in \ai2thor and $1$ meter in \robothor.
Finally, we have to predict the equality predicate, i.e., when two objects
$c$ and $d$ with features $\bz_c$ and $\bz_d$ represent the same object.
To this purpose, we compute the distance
between the two estimated object positions, and assign the object features to the same
object instance whether such a distance is lower than a given
threshold (set to 20 cm in our experiments).
All the training, validation, and testing data have been extracted from
a set of images collected by navigating in the \ai2thor\ simulator. 

\paragraph{Evaluation metrics.}

The evaluation is provided by calculating a number of standard metrics over a set
of episodes. For each task, an episode is obtained by randomly placing
the agent in a random unseen scene and providing it a randomly generated goal for the given task. 
The generated goals are feasible since the object types used in their definitions are randomly chosen from a proper set of types; e.g., the goal to open a box is defined by randomly choosing the type ``box'' from a the set of object types that can be opened.
For all the tasks we adopt the following standard evaluation metrics:
\begin{description}
\item[Success rate (Success):] is equal to the fraction of successful episodes on
  the total number of episodes.
\item[Distance To Success (DTS):] For tasks
  (\textsc{ObjNav} $t_1$), (\textsc{Open} $t_1$), and (\textsc{Close} $t_1$), it is the average
  distance between the agent and the closest object of type $t_1$; for the task (\textsc{On} \,$t_1\, t_2$),
  it is the average distance between the closest pair of objects of types $t_1$ and $t_2$. 
  If the episode succeeds such a distance is set to 0.
\end{description}




In order to measure the impact of errors in object detecting, for each task we consider two versions of \arch. In a version the set of objects $\C$ are those returned by our object detector; in the second version
the set of objects $\C_{GT}$ are those returned by the \ai2thor\ simulator, which corresponds to a ground-truth object detector. 
Moreover, for all tasks, we evaluate the precision $P_{\C}$ and recall $R_{\C}$ of the detected objects, and the precision $\PY$ and recall $\RY$ of their predicate relations. $\PY$ and $\RY$ take into account only the objects that match with ground-truth ones. The matching is performed by computing the Intersection over Union (IoU) among the 2D bounding box detected during the episode and the ground-truth ones: if the IoU is higher than 50\% for a ground-truth object of the same class, then the detected object matches with it.

\paragraph{Experimental results.}
\begin{table}[t]
\centering
\setlength{\tabcolsep}{3pt}
\scalebox{0.74}{
 \begin{tabular}{l c c | c c | c c | c c | c c | c c}
 \hline
  & \multicolumn{2}{c |}{\emph{Success}$\,\uparrow$} & \multicolumn{2}{c |}{\emph{DTS}$\,\downarrow$} & \multicolumn{2}{c |}{$P_{\C}\uparrow$} & \multicolumn{2}{c |}{$R_{\C}\uparrow$} & \multicolumn{2}{c |}{$\PY\uparrow$} & \multicolumn{2}{c }{$\RY\uparrow$} \\ 
  & $\C$ & $\C_{GT}$ & $\C$ & $\C_{GT}$ & $\C$ & $\C_{GT}$ & $\C$ & $\C_{GT}$ & $\C$ & $\C_{GT}$ & $\C$ & $\C_{GT}$ \\ 
 \hline\hline
 \textsc{On} & 0.5 & 0.8 & 1 & 0.37 & 0.28 & 1 & 0.86 & 1 & 0.83 & 0.82 & 0.8 & 0.87 \\ 
 \textsc{Open} & 0.75 & 0.87 & 0.45 & 0.25 & 0.35 & 1 & 0.78 & 1 & 0.82 & 0.81 & 0.72 & 0.82 \\
 \textsc{Close} & 0.78 & 0.89 & 0.39 & 0.16 & 0.32 & 1 & 0.8 & 1 & 0.8 & 0.79 & 0.73 & 0.82 \\
 \textsc{objNav} & 0.78 & 0.83 & 0.27 & 0.19 & 0.42 & 1 & 0.8 & 1 & 0.82 & 0.8 & 0.75 & 0.84 \\
 \hline
 \end{tabular}
 }
 \caption{
 Performance of \arch\ with/out the ground-truth object detection, evaluated on the considered tasks in the \ai2thor simulator. $\uparrow$/$\downarrow$ means the higher/lower the better.
 }
 \label{tab:ithor}
\end{table} In our experiments, a run of \arch\ consists of 200 steps, where at each
step a low-level operation is performed; we call each of these runs an episode.
For all tasks,
the episode dataset uses the {\it test} scenes of \ai2thor, i.e., all environments that does
not appear in the datasets generated for training the predicate
classifiers and object detector. 

In Table \ref{tab:ithor}, we report the average results of all tasks with and
without ground-truth object detection over the considered episodes. For task \textsc{on}, we randomly generated 400 different goals, defining 400 episodes; for tasks \textsc{open} and \textsc{close}, we randomly generated 100 goals, defining 100 episodes for each task; for the object goal navigation task, we used the test set of goals proposed in \cite{wortsman2019learning}, defining 2133 episodes.
It is worth noting
that, for the object goal navigation task,  two different
episodes often have the same goal but a different initial pose of the agent. 

The impact of errors in object detecting for tasks \textsc{objNav}, \textsc{open} and \textsc{close} is pretty low and, as expected, it is the half of the impact for task \textsc{on}, since this latter task requires to detect two objects, while all other tasks requires to detect a single object.
Without ground-truth object detection, \arch\ achieves the best success rate on the object goal navigation task; same or similar results are also provided in tasks \textsc{open} and \textsc{close}, since they can be seen as an extension of the object goal navigation task where, after finding and going near to an object, the agent has only to open or close the object. In the \textsc{on} task, the success rate decreases significantly, because it requires moving towards two objects, instead of only one, and has two additional complexities given by the facts that one object must be placed on the other one in a clear place, i.e., a place not obstructed by other objects, and that the total encumbrance of the agent increases when it carries an object, which causes more collisions during the navigation. 


Metric $P_\C$ measures the amount of false positives in detecting objects. Although the values of $P_\C$ is quite low for almost all the considered tasks, the success rate is relatively high  because (i) many false positive  objects are not involved in the definition of goals, and (ii) the agent acts by using the objects with the highest confidence, which usually correspond to ground-truth objects.
%
%
$P_{\C}$ is higher for the object goal navigation task, because for this task the agent achieves the goal in fewer steps than for other tasks, and this reduces the number of predictions and the chance of detecting false positive objects. 

Metric $R_\C$ measures the amount of true positive detected objects. The values for $R_{\C}$ are quite high, and hence the real existing objects are often detected, although in our experiments the agent sometimes fails to recognize objects when they are far from the agent.
Moreover, the values of $\PY$ and $\RY$ are relatively high, and hence the agent can construct a symbolic state that is quite correct and complete, enabling an effective planning.

As expected, when \arch\ is provided with ground-truth object detection, all metrics are better than or similar to using our object detection. Only $\PY$ is slightly lower when ground-truth object detection is used; we think this is due to the fact that sometimes the ground-truth object detection identifies objects which are only partially seen by the agent camera and predicting their properties more likely fails (e.g., the agent fails in predicting whether a fridge is open when it sees only a corner of the fridge).

\subsection{Comparison on object goal navigation}
\begin{table}[t]
\centering
\scalebox{0.85}
{
 \begin{tabular}{l c c c} 
 \hline
  & \emph{Success}$\,\uparrow$ & \emph{SPL}$\,\uparrow$ \\ 
 \hline\hline
 Random & 1.72\% & 1.33\% \\
 DD-PPO & 35.11\% & 17.37\% \\
 DD-PPO$_{action\_boost}$ & 36.61\% & 17.49\% \\
 \arch & \bf{56.78}\% & \bf{24.87}\%\\
 \hline
 \end{tabular}
 }
 \caption{Performance of \arch\ w.r.t.\ the random baseline, DD-PPO, and DD-PPO$_{action\_boost}$, evaluated on the object goal navigation task in the \robothor\ simulator. 
 }
 \label{tab:robothor}
\end{table}
We did not find other approaches using simulator \ai2thor that solve the tasks considered in our experiments. So, in our experimental analysis we considered a second simulator, \robothor\ \cite{robothor}, for which the last challenge concerning the object goal navigation was launched in 2021. 

For the object goal navigation task, we compared \arch\ with a random baseline, an RL baseline provided in the challenge, called DD-PPO, and the winner of the challenge, called DD-PPO$_{action\_boost}$. 
Both the RL baseline and the winner exploit the DD-PPO algorithm \cite{wijmans2019ddppo} where the hidden state is computed by providing, as input to a GRU \cite{cho2014properties}, the visual features of the RGB-D images computed by a ResNet-18 \cite{he2016deep}. The baseline and the winner approach have been trained on 108,000 episodes for 300 and about 10 million steps, respectively.

For this experiment, we adopt an additional metric, called Success weighted by Path Length (SPL) and introduced by \citeauthor{anderson2018evaluation} (\citeyear{anderson2018evaluation}). This metric measures the efficiency of the agent in reaching the goals and is defined as: $$SPL=\frac{1}{N}\cdot \sum_{i=1}^N \left(s_i\cdot \frac{p_i^\star}{max(p_i, p_i^\star)}\right)$$
where $N$ is the number of episodes, $p_i^\star$ is the shortest-path distance from the initial position of the agent to the closest goal in the $i$-th episode, $p_i$ is the length of the agent path in the $i$-th episode, and $s_i$ is a boolean variable equal to 1 when the $i$-th episode succeeds, and equal to 0 otherwise. If the path of the agent is the shortest one, the term in parenthesis is 1. The longer the path, the lower the term in parenthesis and the worse the metric.

For the experiment, we considered the validation episode dataset provided in the challenge, which is composed by 1800 episodes set in the 15 validation scenes of \robothor. We did not consider the test episode dataset of the challenge, because for such a dataset the evaluation can be done only by the organizers of the challenge who require that the evaluated approach plays by the challenge rule. This is not the case for \arch\ because it allows the agent to perceive its pose, which is not available in the challenge. While the usage of this additional information can in principle favors  \arch\ w.r.t.\ the approaches that took part in the challenge, it is worth noting that the agent position can be approximately derived from the RGB-D egocentric views by means of visual simultaneous localization and mapping methods \cite{taketomi2017visual}. Most importantly, the usage of the validation dataset of \robothor\ disfavors \arch\ w.r.t.\ the other compared approaches because 
the object detector and predicate classifiers of \arch\ are trained using the training and validation scenes of a different simulation environment, \ai2thor, while the other compared approaches are trained and validated on the training and validation scenes of \robothor. 
%

Each episode of the dataset consists of 500 steps, and regards finding and moving toward objects of 12 types. We trained an object detector similarly to the one for \ai2thor simulator, but focused on the 12 goal object types of \robothor, which provides a performance slightly higher than the object detector trained using all the 118 object types of \ai2thor, obtaining 59.02\% precision and 69.06\% recall.

Table \ref{tab:robothor} shows the results of the comparison. The random baseline provides poor performances. 
This indicates that, for the \robothor\ simulator, the object goal navigation task is quite challenging. 
The complexity of the task is confirmed by the performance of the RL baseline which is  higher than the random baseline but still quite low. DD-PPO$_{action\_boost}$ provides results slightly higher than the RL baseline. Remarkably, \arch\ outperforms DD-PPO$_{action\_boost}$ in terms of success rate and \emph{SPL}. This confirms that the integration of symbolic planning with state recognition from sensory data can provide competitive results w.r.t.\ RL based approaches.

\subsection{Error analysis}

 \begin{figure}[t]
\centering
\includegraphics[width=.81\linewidth]{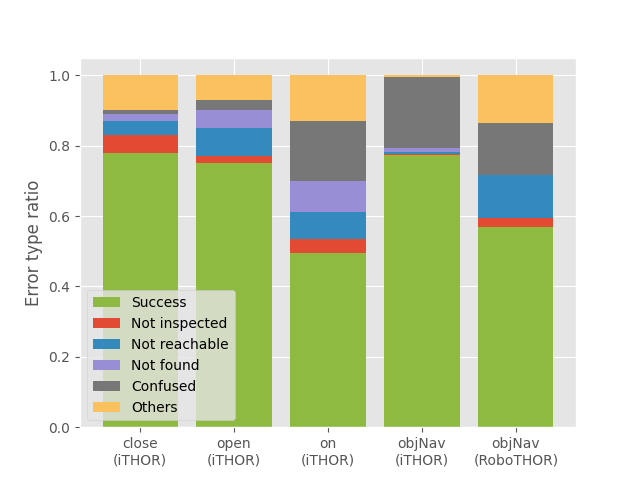}
\caption{Ratio of the occurrences of different error types made by \arch. 
}
\label{fig:error_analysis}
\end{figure}

In Figure \ref{fig:error_analysis}, we analyze the errors made by \arch\ on all tasks. For few episodes, denoted as ``Not inspected'', the agent detects a far object of the same type as the type used for the goal definition, and subsequently approaches the object but is no more able to recognize it. This is due to the fact that either the object does not really exists, or the agent does not recognize an existing  object, despite being close to and looking at it.
For some episodes, namely ``Not reachable'', the agent finds a goal object but cannot reach a position close enough to the object. This can be due to the fact that either the agent collides or the goal object estimated position is farther than the real one.
Collisions more often happen for the task \textsc{on}, when the agent holds an object as the agent encumbrance increases. An error in the estimation of the object position is more likely for large objects, such as tables or televisions, since the agent considers the center of the object as its position. There are few episodes, labelled as ``Not found'', where the agent does not find the object, due to either an ineffective exploration of the environment or false negatives of the object detector. We observed that the latter case is more likely than the former, because the agent almost always explores the entire environment within the given number of steps. The errors labelled by ``Confused'' denote episodes for which the agent believes it succeeded while the task has not been completed. This is due to false positives of the object detector. Finally, ``Others'' denote all other task-dependent failures. 
E.g., for the \textsc{on}, \textsc{open} and \textsc{close} tasks, the agent sometimes fails to identify the object position when it has to manipulate an object. 
This more likely happens for small objects, such as spoons or saltshakers.
Moreover, for the \textsc{on} predicate an agent can fail to put an object on a table due to the fact that the target position is already occupied, or there is not enough space on the table.


 \begin{figure}[t]
\centering
\includegraphics[width=.81\linewidth]{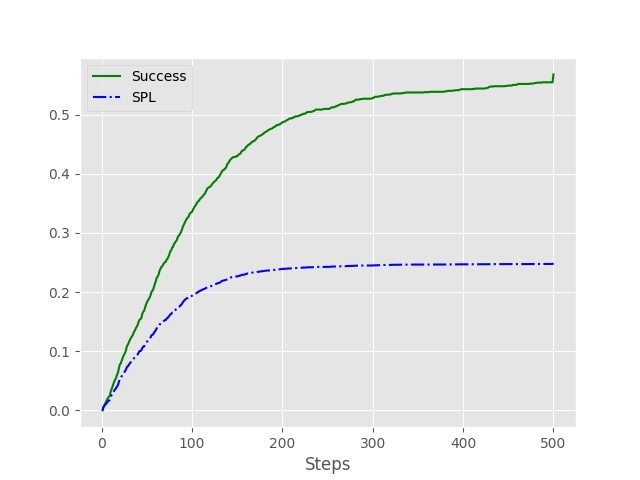}
\caption{Average performance of \arch\ for the goal object navigation task in the \robothor simulator, using a number of steps ranging from 0 to 500. 
}
\label{fig:ogn_metrics_evolution_robothor}
\end{figure}
Figure \ref{fig:ogn_metrics_evolution_robothor} shows the success rate and \emph{SPL} for a number of steps ranging from 0 to 500. For almost all episodes the agent achieves the goal in 300 steps. 
For few episodes, the agent achieves the goal only after 500 steps. This happens because the agent is actually close to and looks at a goal object, but it fails to recognize the object.



\section{Conclusions and Future Work}
We have proposed a framework, called \arch, for the online grounding of planning domains in unknown environments. Our approach enables an agent to map the sensory data into a symbolic state, allowing to perform and exploit efficient planning in a wide variety of different environments. We have tested the proposed method on different tasks obtaining better results than recent RL-based approaches. Future work will focus on learning a policy to compile the high-level actions into low-level executable operations, 
and on learning, online, the mapping of the sensory
representations to symbolic ones.


\bibliographystyle{kr}
\bibliography{ijcai22}

\end{document}